\documentclass[letterpaper, 10 pt, conference]{ieeeconf}

\IEEEoverridecommandlockouts                              % This command is only needed if 
\usepackage{graphicx}
\usepackage{amsmath} % assumes amsmath package installed
\usepackage{amssymb}  % assumes amsmath package installed
\usepackage{xcolor}
\usepackage{xcolor}
\newif\ifhighlightchanges
\highlightchangesfalse % use \highlightchangesfalse for the clean paper
\newcommand{\revised}[1]{%
  \ifhighlightchanges{\color{blue}#1}\else#1\fi}

\newcommand{\tancraft}{TanCRAFT}
\title{\LARGE \bf
\tancraft: Context-Reasoned Anticipatory Feedback for Assembly Tasks
}

\author{
Sassan Mokhtar$^{1}$,
Lars Doorenbos$^{1,2}$,
Marius Bock$^{1,2}$,
Sina Mokhtar$^{3}$,\\[4pt]
Katharina Beckh$^{2,4}$,
Dominik Bach$^{1}$,
and J\"urgen Gall$^{1,2}$%
\thanks{$^{1}$University of Bonn, Germany.}%
\thanks{$^{2}$Lamarr Institute, Germany.}%
\thanks{$^{3}$University of New Mexico, USA.}%
\thanks{$^{4}$Fraunhofer IAIS, Germany.}%
}
\begin{document}

\maketitle
% \thispagestyle{empty}
% \pagestyle{empty}

%%%%%%%%%%%%%%%%%%%%%%%%%%%%%%%%%%%%%%%%%%%%%%%%%%%%%%%%%%%%%%%%%%%%%%%%%%%%%%%%
\begin{abstract}
Preventing human error in physical assembly tasks requires assistance systems that can intervene before mistakes are made. This is especially challenging in settings where users are free to choose the construction order and correctness depends on the evolving assembly rather than a fixed next step.
We present \tancraft{}, a four-stage real-time predictive assistance system that \emph{tracks} the scene state, \emph{reasons} about the current partial assembly, \emph{forecasts} the near-future motion of manipulated pieces, and \emph{intervenes} when the predicted placement is incompatible with the target. 
We instantiate our framework and confirm its effectiveness in a controlled tangram assembly task through offline benchmarks and a within-subject user study. 
Our findings show that the predictive feedback significantly reduces mistakes relative to unguided and reactive sessions, while yielding the lowest mean perceived workload and shortest mean completion time. 
Overall, these results demonstrate the value of predictive feedback for error prevention and provide a step toward cost-aware assistance for automation and human-robot workcells.
\end{abstract}

%%%%%%%%%%%%%%%%%%%%%%%%%%%%%%%%%%%%%%%%%%%%%%%%%%%%%%%%%%%%%%%%%%%%%%%%%%%%%%%%

%%%%%%%%%%%%%%%%%%%%%%%%%%%%%%%%%%%%%%%%%%%%%%%%%%%%%%%%%%%%%%%%%%%%%%%%%%%%%%%%

\section{INTRODUCTION}
\label{sec:introduction}

\revised{Human error remains a significant contributor to failures and quality defects in manual assembly, industrial production, and human-robot workcells~\cite{saptari2015effect,torres2021classification,antonelli2019predicting}.} While some mistakes are easily reversed, others are costly once completed. The cost therefore depends not only on \textit{whether} errors are detected, but also on \textit{when}. For example, a misplaced component or unsafe action may be easy to correct before commitment, but costly, time-consuming, or even unsafe after completion. This motivates the need for assistance systems that go beyond post-hoc error detection and are capable of real-time error forecasting and prevention. 

Anticipating errors is challenging because systems must infer human actions, intentions, and potential mistakes from external observations. Moreover, anticipation alone is insufficient for assistance, as it does not address how a system should act to prevent the mistake. 
\revised{In this context, previous predictive-assistance methods anticipate supportive robot behaviors or human assembly actions~\cite{grigore2018preference, nemlekar2023transfer}. TRAFA~\cite{mokhtar2026trafaanticipatinguseractions} extends prediction to proactive error feedback but operates in a sequential task with a fixed next correct action.} 
%Previous predictive-assistance methods address this by anticipating supportive robot behaviors or human assembly actions, but operate in sequential tasks with fixed next correct actions~\cite{grigore2018preference, nemlekar2023transfer,mokhtar2026trafaanticipatinguseractions}. 
Still, many real scenarios, such as kit or furniture assembly, are far less constrained, letting users choose their own strategy, revise placements, and manipulate pieces without committing them. This freedom makes interaction more natural, but error anticipation substantially harder. %, since correctness must in large part be inferred from the evolving workspace. 
To address this challenge, we propose \tancraft{}, a real-time predictive assistance framework for non-sequential assembly comprising four stages: \emph{Track}, \emph{Reason}, \emph{Forecast}, and \emph{Intervene}. 
% The system tracks the scene state, reasons about the partial assembly, forecasts the near-future motion of manipulated pieces, and intervenes through audio-visual feedback when a predicted placement is incompatible with the target.  
% In this way, \tancraft{} treats feedback as a selective intervention problem, warning not because a piece is touched or motion is uncertain, but because the predicted outcome is likely to lead to an error in the assembly.

\begin{figure}[t]
\centering
\includegraphics[width=\columnwidth]{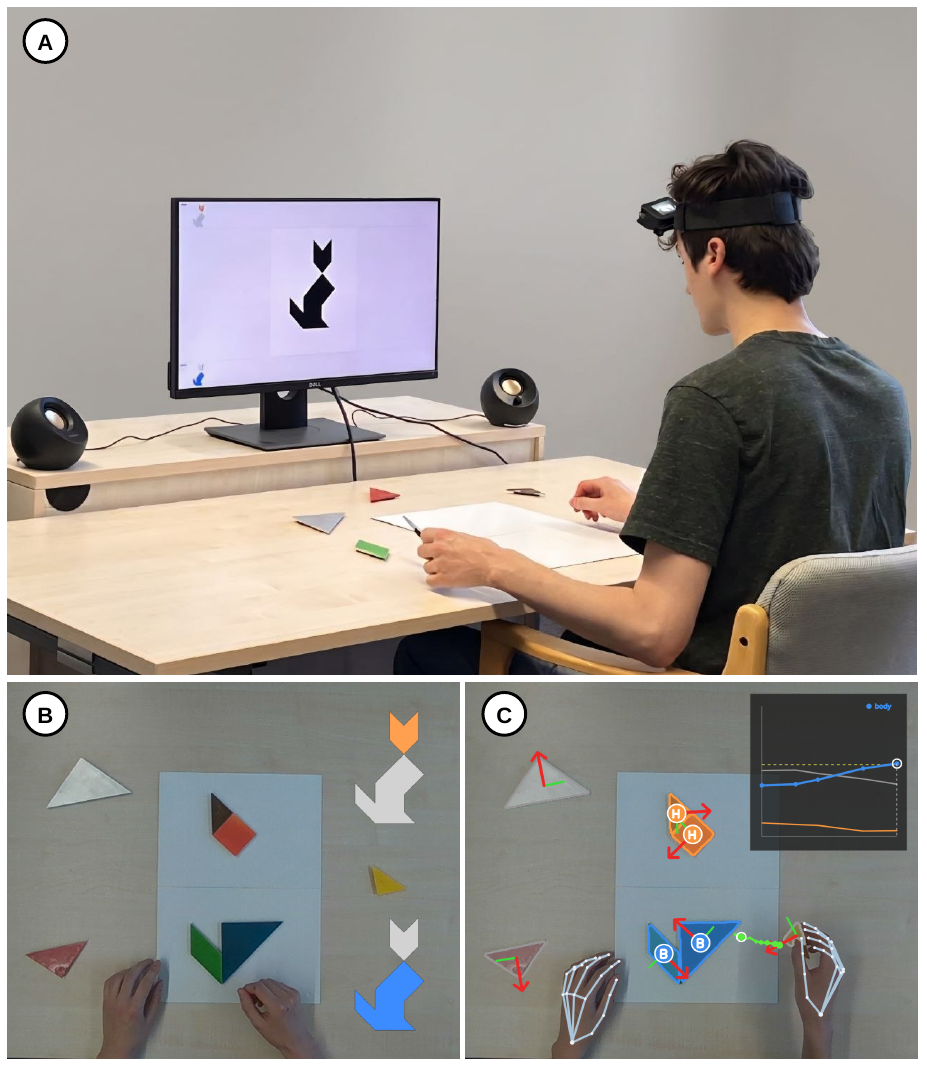}
\caption{\textbf{\tancraft{} overview.} (A) The user assembles the target tangram shape while the system observes the workspace. (B) The system reasons about the evolving assembly state and assigns placed pieces to target-shape subgraphs. (C) As the user moves a piece toward an incompatible location, \tancraft{} 
% forecasts its motion, estimates the intended destination, and 
issues audio-visual feedback to prevent the mistake.}
% when the invalid-placement confidence exceeds the intervention threshold.}
\label{fig:teaser}
\end{figure}

We instantiate the framework in a controlled tangram assembly task, which recently gained popularity in robotics assembly and manipulation tasks because it combines objective correctness with an open construction order, interchangeable pieces, and exploratory manipulation~\cite{zhao2025master, lee2025imagine}, i.e., the same properties that make unconstrained assembly a difficult task. 
Unlike prior works, where robots assemble the pieces, we use tangram to study predictive assistance to humans, as it provides a natural way to induce human error without the costs associated with a full industrial setup.
% validate the reasoning, forecasting, and intent-estimation modules, while a system-level comparison evaluates mistake anticipation against four learned baselines. 
We validate our system through offline benchmarks, and evaluate the full system in a within-subject user study comparing two variants of \tancraft{} to unguided and reactive (post-hoc) feedback conditions. 
The user study shows that predictive feedback significantly outperforms both no-feedback and reactive feedback baselines in mistake reduction, while also achieving the lowest mean perceived workload~\cite{hart1988development}, the highest success rate, and the shortest mean completion time.

In short, the main contributions of this work are:
\begin{itemize}
    \item We introduce \tancraft{}, a real-time framework for predictive assistance in non-sequential assembly tasks.
    % , and demonstrate improved mistake anticipation over learned baselines, including on held-out target shapes.
    \item We show through a user study that predictive feedback significantly reduces mistakes, while achieving the lowest mean workload and shortest mean completion time.
     % relative to no feedback and reactive feedback
    \item \revised{We will release the code, trained model weights, and multi-view tangram assembly dataset}
    % with approximately $16$ hours of synchronized overhead/egocentric RGB videos, tracked objects, hand states, placements, mistakes, and feedback metadata 
    upon acceptance.
\end{itemize}

%%%%%%%%%%%%%%%%%%%%%%%%%%%%%%%%%%%%%%%%%%%%%%%%%%%%%%%%%%%%%%%%%%%%%%%%%%%%%%%%
\section{RELATED WORK}

\subsection{Predictive assistance for robots in HRC and assembly}
\label{sec:predictive_assistance}

Predictive assistance has been widely studied in human-robot collaboration (HRC), particularly in assembly and manufacturing, where robots need to anticipate human behavior to adapt their actions. 
Prior work has modeled multi-step activity for robot planning~\cite{hawkins2013probabilistic}, introduced hierarchical planning and task allocation for collaborative 
assembly~\cite{johannsmeier2016hierarchical}, and integrated human motion prediction with robot planning~\cite{unhelkar2018human}. 
Other approaches combine plan recognition and trajectory prediction to infer human plans and adapt robot actions accordingly~\cite{cheng2020towards}. \revised{Preference-based methods predict supportive robot behaviors~\cite{grigore2018preference} or adapt to human assembly preferences for proactive part delivery~\cite{nemlekar2023transfer}. 
In contrast, our focus is on anticipating human errors during ongoing human manipulation, where exploratory motion makes placement intent ambiguous, and feedback must arrive early enough to redirect the piece.}
% This requires combining motion forecasting with reasoning about the evolving assembly state to distinguish valid exploration from an impending mistake.

\subsection{Feedback, guidance, and the cost of intervention}
\label{sec:feedback_cost}

Feedback and guidance are commonly used in assisted manual work to communicate task instructions, support execution, or flag deviations from the intended procedure~\cite{eversberg2023evaluating,marino2024assessing}. However, online feedback during an ongoing action also competes for the user's attention. Interruption studies show that unsolicited messages can increase completion time, error rate, annoyance, and anxiety, with a cost that depends on the user's task state when they are delivered~\cite{bailey2006need,adamczyk2004if}. Warning-system research further shows that unreliable alerts and false alarms can reduce compliance and reliance on automated warnings~\cite{dixon2007independence}. \revised{In human-robot collaborations, Mannem et al.~\cite{mannem2023exploring} show that robot queries can reduce perceived helpfulness even when objective task performance is largely unaffected. LeMasurier et al.~\cite{lemasurier2024reactive} anticipate robot failures and provide proactive explanations, which participants rate more favorably than reactive explanations. Together with work on cost-aware suggestion timing~\cite{yu2022optimizing}, these studies motivate selective, cost-aware intervention in non-sequential assembly: feedback should help prevent errors while avoiding unnecessary interruptions and additional workload.}

\subsection{Predictive feedback beyond sequential tasks}
\label{sec:predictive_sequential}

Predictive feedback for human users in non-sequential assembly remains largely unexplored. 
%Most closely related to our work is 
TRAFA~\cite{mokhtar2026trafaanticipatinguseractions} uses a Track-Forecast-Act loop to warn users before incorrect actions are completed. However, TRAFA operates in a sequential Duplo-stacking task with a fixed target order. In contrast, tangram assembly allows users to choose their placement order and construct multiple partial structures in parallel, making fixed next-step rules insufficient. \revised{\tancraft{} addresses this challenge by combining motion forecasting with reasoning about the evolving assembly: it predicts where a piece will be placed and checks whether that subgraph can still accommodate its type. This allows the system to anticipate errors without assuming a fixed placement order.}

% We study whether this assembly-aware feedback reduces mistakes in open-order assembly without imposing the workload associated with continuous feedback.

%%%%%%%%%%%%%%%%%%%%%%%%%%%%%%%%%%%%%%%%%%%%%%%%%%%%%%%%%%%%%%%%%%%%%%%%%%%%%%%%
\begin{figure}[t]
\centering
\includegraphics[width=0.9\columnwidth]{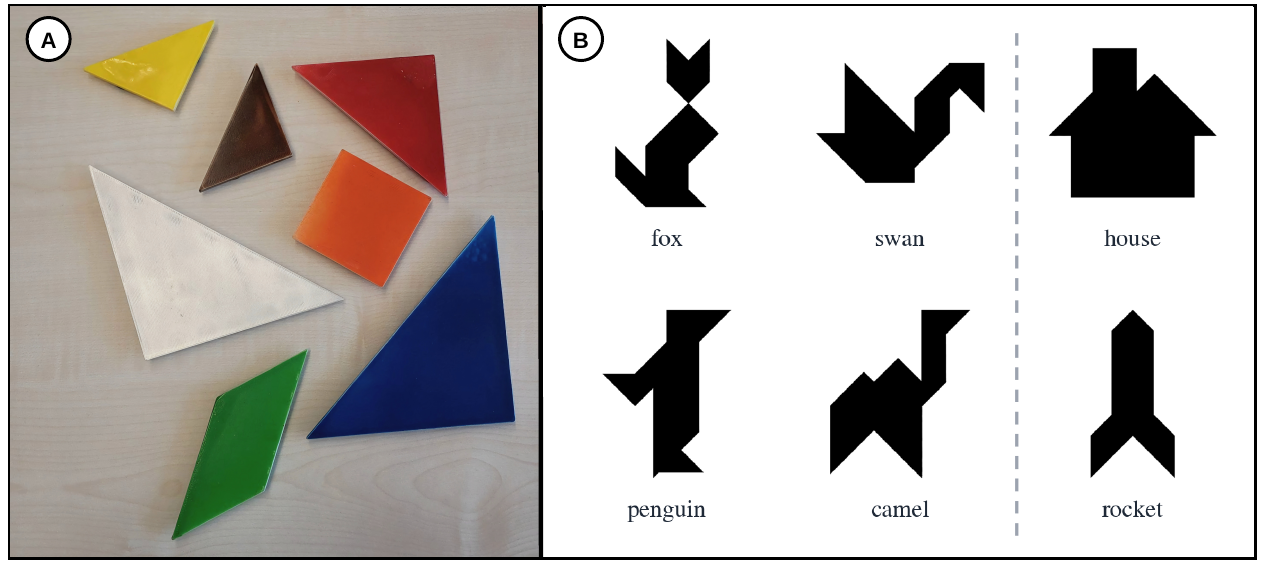}
\caption{\textbf{Tangram pieces and target shapes used in the study.} 
(A) The seven physical tangram pieces used for assembly. (B) The six target shapes used during data collection and training. In the user study, the rocket was used in the written instructions, the house was used for the warm-up session, and the four animal shapes were used for testing the feedback modes.}
\label{fig:tang_piece}
\end{figure}
%%%%%%%%%%%%%%%%%%%%%%%%%%%%%%%%%%%%%%%%%%%%%%%%%%%%%%%%%%%%%%%%%%%%%%%%%%%%%%%%

\section{METHOD}
\label{sec:method}

Users recreate a target shape $S$ from the Tangram pieces $\mathcal{P}=\{p_i\}_{i=1}^N$, where each piece has a geometric type $\tau(p_i)\in\mathcal{T}$. The pieces consist of two large triangles, one medium triangle, two small triangles, one square, and one parallelogram, with same-type pieces interchangeable (Fig.~\ref{fig:tang_piece}). A completed assembly is correct when the final configuration of all pieces is geometrically consistent with the target shape $S$. We decompose the target shape into subgraphs $\mathcal{G}(S)=\{g_k\}_{k=1}^K$, where each subgraph corresponds to a part of the target shape, such as the head or body of an animal. Each subgraph has a prescribed composition, where $r_k(\tau)$ denotes the required number of pieces of type $\tau$ in subgraph $g_k$.
 
We restrict our notion of a mistake to consequential errors: placing a piece into a subgraph with no remaining capacity of its type. This excludes minor adjustments made during placement, such as when orienting a piece within its intended subgraph. 
We denote the set of pieces being manipulated at time $t$ by $\mathcal{A}_t \subseteq \mathcal{P}$. Then, for an active piece $a_t \in \mathcal{A}_t$, let $n_k^t(\tau)$ count the other pieces of type $\tau$ currently assigned to subgraph $g_k$. A subgraph is invalid for an active piece $a_t$ when the subgraph has no remaining capacity for that piece's type, i.e., $n_k^t(\tau(a_t)) \geq r_k(\tau(a_t))$. This, in turn, forms the set of invalid subgraph destinations
\begin{equation}
\label{eq:mistake}
    \mathcal{I}_t(a_t) = \{\, g_k \in \mathcal{G}(S) : n_k^t(\tau(a_t)) \geq r_k(\tau(a_t)) \,\}.
\end{equation}
This set changes as pieces are placed, removed, or reassigned. Thus, our task is to anticipate, for each active piece and before its placement is complete, whether it is being moved into a subgraph in $\mathcal{I}_t(a_t)$, and to intervene only in that case.

%-----------------------------------------------------------------------------
\begin{figure*}[t]
\centering
\includegraphics[width=\textwidth]{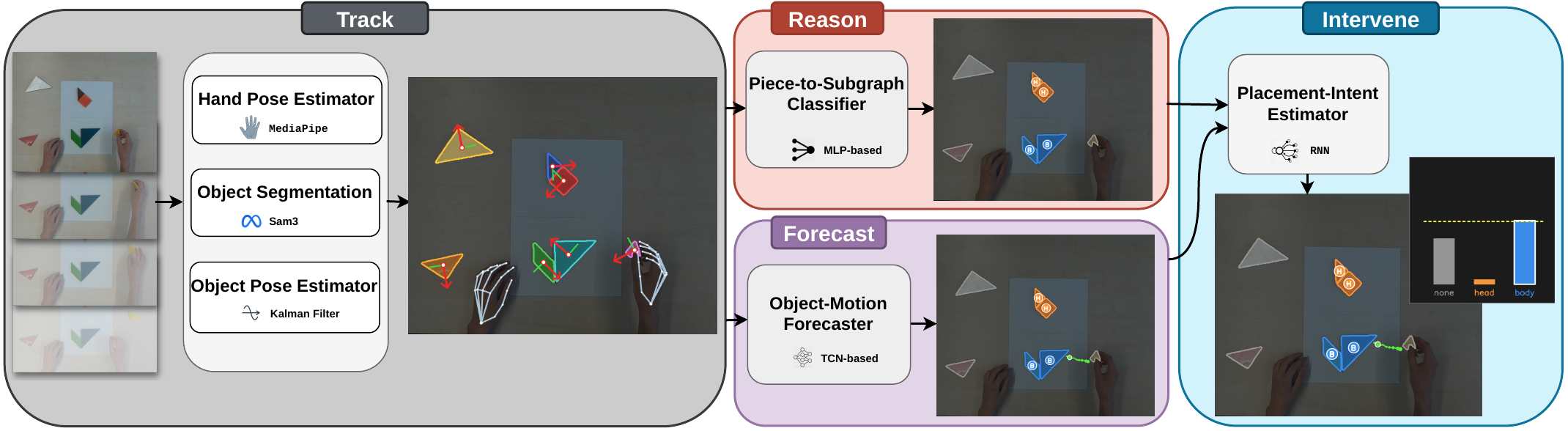}
\caption{\textbf{Overview of \tancraft{}.}
Given an overhead RGB stream, the 
\textit{Track} stage estimates hand poses, tracks tangram-piece masks, and estimates object poses using a Kalman filter. The \textit{Reason} stage infers the evolving assembly state with a Piece-to-Subgraph Classifier. The \textit{Forecast} stage predicts the near-future motion of active pieces from their recent trajectories and scene context. Finally, the \textit{Intervene} stage estimates active-piece destinations and checks their compatibility with the remaining subgraph capacities to drive predictive audio-visual warnings or continuous feedback.}
\label{fig:pipeline}
\end{figure*}
%-----------------------------------------------------------------------------

\subsection{System Overview}
\label{sec:system_overview}

Our framework forecasts the mistakes defined above from an overhead RGB stream and converts them into actionable feedback. At time \(t\), the system processes the overhead RGB stream to obtain and maintain a task state
\begin{equation}
    \mathbf{x}_t =
    \left(
    \{\mathbf{z}_i^t\}_{i=1}^{N},
    \mathbf{h}_t,
    \{\hat{s}_i^t\}_{i=1}^{N},
    \mathcal{A}_t,
    S
    \right),
\end{equation}
where $\mathbf{z}_i^t$ denotes the estimated planar position and orientation of piece \(p_i\), \(\mathbf{h}_t\) denotes the estimated hand state, and $\hat{s}_i^t \in \mathcal{G}(S)\cup\{\varnothing\}$ is the predicted subgraph assignment of piece \(p_i\), with \(\varnothing\) indicating that a piece is not currently assigned to any target subgraph. The system uses this state to anticipate incompatible placements through four stages: \emph{Track}, \emph{Reason}, \emph{Forecast}, and \emph{Intervene} (Fig.~\ref{fig:pipeline}).

%-----------------------------------------------------------------------------

\subsection{Track: Perception and Scene-State Estimation}
\label{sec:perception}

The first stage converts the overhead RGB stream into the scene state by segmenting tangram pieces, assigning persistent instance identities, deriving the piece type from mask geometry, and using color to disambiguate shape duplicates.
We use SAM3~\cite{carion2025sam} to initialize and propagate piece masks through the video stream. 
% To suppress identity drift between touching pieces, we keep the clean initialization frame in SAM3's memory and constrain the tracked masks to be non-overlapping. 
From each tracked mask, we extract the image-plane center and in-plane orientation, yielding the planar pose,
\begin{equation}
\mathbf{z}_i^t = (\mathbf{c}_i^t,\theta_i^t),
\end{equation}
where \(\mathbf{c}_i^t \in \mathbb{R}^2\) is the piece center and \(\theta_i^t\in[0,2\pi)\) is its orientation.
In parallel, we use MediaPipe Hands~\cite{lugaresi2019mediapipe} to estimate 2D hand keypoints, which define the hand pose $\mathbf{h}_t$.

A key difficulty is hand occlusion: the active piece is often covered when its motion is most informative for forecasting. We therefore track each piece with a constant-velocity Kalman filter~\cite{kalman1960new} over its planar pose, using measurement updates when reliable masks are available and prediction-only updates otherwise. 
This smoothes trajectories and bridges short gaps in case of occlusions. From these, the active object set $\mathcal{A}_t$ is inferred from hand–object contact and recent object motion, denoting a piece as active when it overlaps with the fingertips. %  coincides with consistent displacement

%-----------------------------------------------------------------------------

\subsection{Reason: Unconstrained Assembly Reasoning}
\label{sec:assembly_reasoning}

The \textit{Reason} stage estimates the current assembly state by assigning each piece \(p_i\) to a subgraph in $\mathcal{G}(S)$ or $\varnothing$. The assignments of all pieces are denoted by $\hat{\mathbf{s}}^t=(\hat{s}_1^t,\ldots,\hat{s}_N^t)$. For the assignment, we use a learned Piece-to-Subgraph Classifier $f_\phi$, applied independently to each piece. We first group spatially adjacent pieces into clusters, which are connected components of the piece-adjacency graph, with adjacency determined by mask dilation and a centroid-distance threshold. The vector $\mathbf{r}_i^t$ encodes which detected cluster contains piece $p_i$, including a separate category for pieces that belong to no clusters. These geometric groupings do not carry the information about the subgraphs, which is inferred by the classifier $f_\phi$. 

Given a piece's planar pose $\mathbf{z}_i^t$, type $\tau(p_i)$, cluster membership \(\mathbf{r}_i^t\), and the target shape $S$, the classifier $f_\phi$ predicts the subgraph in which the piece is placed, or the null class otherwise:
\begin{equation}\label{reason_network}
\boldsymbol{\pi}_i^t =
f_{\phi}
\left(
\mathbf{z}_i^t,
\tau(p_i),
\mathbf{r}_i^t,
S
\right),
\qquad
\boldsymbol{\pi}_i^t \in \Delta^{K+1},
\end{equation}
where $\boldsymbol{\pi}_i^t$ is the probability distribution over the \(K\) subgraphs and the null class, and $\Delta^{K+1}$ denotes the probability simplex. The piece is then assigned to the most likely class,
\begin{equation}
\hat{s}_i^t = \arg\max \boldsymbol{\pi}_i^t.
\end{equation}

The resulting assignments $\hat{\mathbf{s}}^t$ determine the occupancy counts $n_k^t(\tau)$ used in Eq.~\eqref{eq:mistake}. The target shape \(S\) is supplied as a one-hot label vector over the target shapes, which lets a single model operate across all shapes.

%To avoid instability, a piece is committed to a subgraph only once it is no longer active and spatially stable, and the assignment is retained until the piece is manipulated again. 

%-----------------------------------------------------------------------------

\subsection{Forecast: Active-Piece Motion Forecasting}
\label{sec:pose_forecasting}

\revised{The \textit{Forecast} stage uses a learned forecaster $f_\psi$ to predict the future 2D center trajectory of active pieces over $H$ time steps, conditioned on their preceding $T$ observed positions, the current scene context, and formed assembly clusters:}
\begin{equation}\label{eq:forecast}
\hat{\mathbf{c}}_{a_t}^{t+1:t+H}
=
f_{\psi}
\left(
\mathbf{c}_{a_t}^{t-T+1:t},
\{\mathbf{c}_i^{t}\}_{i=1}^{N},
a_t,
\{\bar{\mathbf{c}}_m^{t}\}_{m=1}^{M}
\right),
\end{equation}
where \(\mathbf{c}_{a_t}^{t-T+1:t}\) is the recent position history of the active piece, \(\{\mathbf{c}_i^{t}\}_{i=1}^{N}\) are the current positions of all pieces in the workspace, \(a_t\) is the active piece, and \(\{\bar{\mathbf{c}}_m^{t}\}_{m=1}^{M}\) are the centroids of the \(M\) currently formed assembly clusters.

%We implement $f_{\psi}$ as a temporal convolutional network (TCN)~\cite{farha2019ms}: a TCN encoder summarizes the recent active-piece motion together with the scene context, and a TCN decoder predicts the future trajectory as displacements $\Delta \hat{\mathbf{c}}_{a_t}^{t+1:t+H}$. These displacements are accumulated from the current active-piece position to obtain the future locations:
The forecaster predicts the future trajectory as displacements $\Delta \hat{\mathbf{c}}_{a_t}^{t+1:t+H}$. These displacements are accumulated from the current active-piece position to obtain the future locations:
\begin{equation}
\hat{\mathbf{c}}_{a_t}^{t+\ell}
=
\mathbf{c}_{a_t}^{t}
+
\sum_{j=1}^{\ell}
\Delta \hat{\mathbf{c}}_{a_t}^{t+j},
\qquad
\ell=1,\ldots,H .
\end{equation}
These forecasts are passed to the intervention stage to estimate the active piece's likely destination before placement is complete.
%-----------------------------------------------------------------------------

\subsection{Intervene: Placement-Intent Estimation and Predictive Intervention}
\label{sec:intervention}

The \textit{Intervene} stage estimates whether the ongoing manipulation will produce an invalid placement and decides whether this estimate warrants feedback. It draws on the assembly estimate $\hat{\mathbf{s}}^t$ from the reasoning stage and the forecast trajectory $\hat{\mathbf{c}}_{a_t}^{t+1:t+H}$ from the forecasting stage to estimate the intended destination of the active piece before placement is complete, using a learned placement-intent model \(f_\omega\) that outputs a distribution over possible subgraph destinations:
\begin{equation}\label{eq:intervene}
\mathbf{q}_t =
f_\omega
\left(
\hat{\mathbf{c}}_{a_t}^{t+1:t+H},
\boldsymbol{\delta}_t
\right),
\qquad
\mathbf{q}_t \in \Delta^{K+1},
\end{equation}
where the additional null class represents non-placement or uncertain motion. \revised{The proximity vector $\boldsymbol{\delta}_t\in\mathbb{R}^K$ has entries $\delta_{t,k}=\exp(-d_{t,k}/\sigma)$, where $d_{t,k}$ is the image-plane distance from the active piece's current center to the nearest pixel of subgraph $g_k$'s mask and $\sigma=80$ pixels is a distance scale. For an empty subgraph, $\delta_{t,k}=0$.}

From these predictions, the intervention policy evaluates whether the most likely destination is compatible with the assembly constraints. Specifically, we use the subgraph index and confidence,
\begin{equation}
\hat{k}_t = \arg\max_{k\in\{1,\ldots,K\}} q_{t,k},
\qquad
\hat{q}_t = q_{t,\hat{k}_t},
\end{equation}
to obtain the frame-level invalid-placement confidence
\begin{equation}
\rho_t =
\begin{cases}
\hat{q}_t, & g_{\hat{k}_t}\in \mathcal{I}_t(a_t),\\
0, & \text{otherwise}.
\end{cases}
\end{equation}
The confidence $\rho_t$ corresponds to the predicted probability $\hat{q}_t$ that the active piece $a_t$ will be added to a subgraph $\mathcal{I}_t(a_t)$ \eqref{eq:mistake} it does not belong to. The set $\mathcal{I}_t(a_t)$ changes as pieces are placed, removed, or reassigned. 
%Thus, our task is to anticipate, for each active piece and before its placement is complete, whether it is being moved into a subgraph in $\mathcal{I}_t(a_t)$, and to intervene only in that case. 

Our predictive approach emits a warning according to
\begin{equation}
u_t^{\mathrm{pred}} =
\begin{cases}
\text{warn}, & \rho_t \geq \gamma,\\
\text{silent}, & \rho_t < \gamma,
\end{cases}
\end{equation}
where \(\gamma\) controls the tradeoff between early sensitivity and false alarms. The null class is excluded from destination selection: a warning requires a non-null destination probability of at least $\gamma$.

%-----------------------------------------------------------------------------
% Loss formulas can be cut in the case of page limit issue
\subsection{Training and Online Inference}
\label{sec:training}

% We train the modules offline using automatically tracked object trajectories and sparse manual subgraph annotations from a separate pilot cohort. 
\revised{We train the modules offline using data from a separate pilot cohort, described in Sec.~\ref{sec:Exp_Res}.} The piece-to-subgraph network \(f_{\phi}\) \eqref{reason_network} is a \revised{three-layer multilayer perceptron (MLP)}, trained with cross-entropy over the $K$ target subgraphs and the null class,
\begin{equation}
\mathcal{L}_{\text{subgraph}}
= -\sum_{k=0}^{K} y_{i,k}^t \log \pi_{i,k}^t,
\end{equation}
with $\mathbf{y}_i^t$ being the ground-truth label of the subgraph the piece $p_i$ belongs to at time $t$.

The active-piece forecaster \(f_{\psi}\) \eqref{eq:forecast} is implemented as a temporal convolutional network (TCN)~\cite{farha2019ms}. It encodes three inputs in parallel: the active piece's past trajectory via a two-layer TCN, and the other pieces' positions and the cluster centroids each via a two-layer MLP. These are concatenated into a single fused representation, which a three-layer temporal convolutional decoder maps to per-step displacements. The forecaster is trained with a position regression loss over a 
%history $T$ and 
horizon $H$:
\begin{equation}
\mathcal{L}_{\text{forecast}}
=
\sum_{\ell=1}^{H}
\left\|
\hat{\mathbf{c}}_{a_t}^{t+\ell}
-
\mathbf{c}_{a_t}^{t+\ell}
\right\|_2^2 .
\end{equation}

The placement-intent model \(f_{\omega}\) \eqref{eq:intervene} predicts the destination subgraph of an active piece within its horizon. It encodes the forecast trajectory using a single-layer \revised{gated recurrent unit (GRU)}, concatenates the resulting representation with the proximity vector $\boldsymbol{\delta}_t$, and classifies the features with a three-layer MLP. 
\revised{The model is trained with class-weighted cross-entropy over the \(K+1\) destination classes:
\begin{equation}
\mathcal{L}_{\text{intent}}
= -\sum_{k=0}^{K} w_k y_{t,k}\log q_{t,k},
\end{equation}
where $w_k$ are inverse training-class frequencies, normalized to one.} During online inference, perception updates piece and hand poses and identifies active pieces, while reasoning assigns non-active pieces to target subgraphs. 
For each active piece, the forecaster predicts its short-horizon trajectory, the placement-intent model estimates its destination, and the intervention policy combines destination confidence with subgraph-capacity constraints to trigger predictive warnings or continuous feedback. 
Our system operates at ${\approx} 9.7$ fps, with runtime dominated by SAM3 mask propagation.

%%%%%%%%%%%%%%%%%%%%%%%%%%%%%%%%%%%%%%%%%%%%%%%%%%%%%%%%%%%%%%%%%%%%%%%%%%%%%%%%
\section{Experiments and Results}
\label{sec:Exp_Res}

We evaluate our framework with offline benchmarks (Secs.~\ref{sec:module_benchmarks},~\ref{sec:system_benchmark}) and a user study (Secs.~\ref{sec:experimental_design},~\ref{sec:user_study_results}).
We implement our framework for the standard tangram set, which has $N=7$ pieces, and set $K=2$, decomposing the target shapes into two subgraphs of three and four pieces (e.g., the head and body of an animal). Still, the formulation is general and extends to more pieces and finer subgraph decompositions.

\revised{\subsection{Data Collection and Training Splits}
\label{sec:data_splits}

For developing the modules and creating the offline benchmarks, we collected 72 pilot assembly sessions from 16 participants across six target shapes, with 12 sessions per shape. 
Of these, the first 36 sessions were conducted without feedback, providing the data to train the initial system. Then, 36 sessions used this initial system, with 12 each under reactive (warning after incorrect placement), predictive (warning before anticipated incorrect placement), and continuous feedback (a sustained tone whose loudness reflects predicted error confidence). Pilot trials ended upon completion or after five minutes. Participants assembled the target shape displayed on a monitor; mistakes arose naturally. Following this data collection, we trained and evaluated the final modules on appropriate subsets of these recordings covering the different feedback modalities with both correct and incorrect placements.

The subgraph classifier and motion forecaster use a shape-stratified partition of 60 development and 12 test sessions, with 85\% of annotated frames in the development partition used for training and the remaining 15\% for validation. 
All baselines are trained on the same 60 sessions for developing and 12 for testing.
In contrast, the placement-intent estimator uses only the 48 no-feedback and reactive sessions, excluding predictive feedback modes as these alter the intended placement, yielding 30 training, 6 validation, and 12 test sessions. 
For the full system-level benchmark, we use the 40 no-feedback and reactive sessions from the final user study (Sec.~\ref{sec:experimental_design}), again avoiding trajectories altered by predictive or continuous feedback.}

\subsection{Offline Module Benchmarks}
\label{sec:module_benchmarks}
\revised{Here, we evaluate the three learned modules on the pilot partitions described in Sec.~\ref{sec:data_splits} to validate their design and select configurations for online use.} 

First, we report the accuracy of the piece-to-subgraph classifier in Tab.~\ref{tab:subgraph_benchmark} against two baselines: \emph{Location Prior}, which assigns pieces using training set location priors, and \emph{Bayesian Location Update}, which treats the location distribution as a prior and updates it using a likelihood score based on how well the piece types within a cluster match the expected composition of each subgraph. \revised{The classifier uses 958 training, 168 validation, and 263 test labeled frames.} It achieves \(95.8\%\) overall accuracy (Tab.~\ref{tab:subgraph_benchmark}).

\begin{table}[t]
\centering
\caption{Piece-to-Subgraph classification accuracy.}
\label{tab:subgraph_benchmark}
\small
\setlength{\tabcolsep}{7pt}
\begin{tabular}{lc}
\hline
\textbf{Method} & \textbf{Overall accuracy} \\
\hline
Location Prior & \(71.0\%\) \\
Bayesian Location Update & \(73.4\%\) \\
Piece-to-Subgraph Classifier & \(\mathbf{95.8\%}\) \\
\hline
\end{tabular}
\end{table}

\revised{Second, Tab.~\ref{tab:pose_forecast_benchmark} evaluates active-piece trajectory forecasting using $T=9$ observed positions to predict $H=9$ future positions at approximately 9 Hz, corresponding to roughly one second of history and prediction. The forecasting benchmark uses 26,208 training, 4,718 validation, and 7,513 test trajectory windows. We compare linear extrapolation with input ablations of the same TCN architecture. Additional inputs comprise \emph{Context}, the other pieces' positions; \emph{Cluster}, assembly-cluster centroids; and \emph{Hand}, MediaPipe hand keypoints encoded by a spatial-temporal graph convolutional network (ST-GCN)~\cite{yan2018spatial}.} All learned variants outperform linear extrapolation. Combining piece history, context, and cluster centroids achieves the lowest errors and is selected for deployment.%; adding hand features does not improve this configuration.

\begin{table}[t]
\centering
\caption{Active-piece trajectory forecasting benchmark.}
\label{tab:pose_forecast_benchmark}
\footnotesize
\setlength{\tabcolsep}{3.2pt}
\providecommand{\cmark}{\(\checkmark\)}
\begin{tabular}{lcccccc}
\hline
\textbf{Method} & \textbf{Hand} & \textbf{Context} & \textbf{Cluster} &
\textbf{ADE \(\downarrow\)} & \textbf{FDE \(\downarrow\)} & \textbf{Miss \(\downarrow\)} \\
\hline
Linear Extrap. &  &  &  & 82.0 & 129.9 & 69.6 \\
Proposed &  &  &  & 32.6 & 44.6 & 25.2 \\
Proposed & \cmark &  &  & 33.2 & 45.7 & 24.2 \\
Proposed &  & \cmark &  & 31.7 & 44.8 & 25.2 \\
Proposed &  &  & \cmark & 34.2 & 53.4 & 32.5 \\
Proposed & \cmark & \cmark &  & 33.1 & 45.2 & 24.8 \\
Proposed & \cmark &  & \cmark & 33.7 & 44.9 & 24.6 \\
Proposed &  & \cmark & \cmark & \textbf{31.6} & \textbf{43.0} & \textbf{22.9} \\
Proposed & \cmark & \cmark & \cmark & 33.3 & 45.0 & 25.5 \\
\hline
\end{tabular}

\vspace{2pt}
\footnotesize
All variants use pose history; checkmarks indicate additional inputs. Errors are reported in pixels. ADE is the mean L2 error, FDE is the L2 error at the final frame, and Miss is the percentage of trajectories with \(\mathrm{FDE}>50\).
\end{table}

\revised{Third, Tab.~\ref{tab:intent_benchmark} evaluates destination prediction using 14,626 training, 3,092 validation, and 5,478 test samples, with test performance scored over 641 interaction events.} We compare to a rule-based baseline that predicts a destination when the forecast trajectory enters a subgraph mask, along with ablated versions of our learned module. For evaluation, each interaction event is assigned the first non-null destination whose probability reaches $\gamma$. If the threshold is not reached, we use the most likely class at the last frame the piece was still active. Predictions that retain the piece's previous subgraph are treated as null. For our module, we use $\gamma{=}0.55$. We evaluate the impact of $\gamma$ in Sec.~\ref{sec:system_benchmark}.
Overall, combining both forecast and distance features achieves the highest $F_1$ (0.67), supporting the configuration used in deployment.

\begin{table}[t]
\centering
\caption{Placement-intent benchmark.}
\label{tab:intent_benchmark}
\small
\setlength{\tabcolsep}{4.5pt}
\begin{tabular}{lccc}
\hline
\textbf{Method} & \textbf{F1} & \textbf{Prec.} & \textbf{Recall} \\
\hline
Rule-Based Intent Baseline & 0.50 & 0.49 & 0.51 \\
Distance Features          & 0.58 & 0.50 & 0.69 \\
Forecast Features          & 0.63 & 0.52 & 0.79 \\
Forecast + Distance        & \textbf{0.67} & \textbf{0.58} & \textbf{0.81} \\
\hline
\end{tabular}

\vspace{2pt}
\footnotesize
% All rows are evaluated at the highest-confidence threshold whose mean lead time exceeds four frames.
\end{table}

% Continuous feedback instead increases tone loudness with invalid-placement confidence, saturating at 0.65.

%-----------------------------------------------------------------------------

\begin{figure}[t]
\centering
\includegraphics[width=0.8\columnwidth]{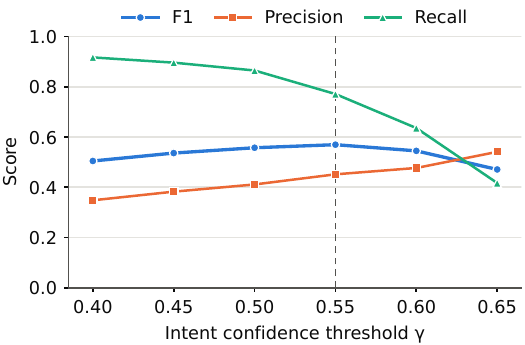}
\caption{\textbf{\revised{System-level precision, recall, and $F_1$ across thresholds.}} The dashed line marks the operating point selected before the user study.}
\label{fig:system_threshold}
\end{figure}

\revised{\subsection{System-Level Mistake Anticipation}
\label{sec:system_benchmark}

After validating the design of the individual modules, we now turn to evaluating the complete reasoning, forecasting, and intervention pipeline against other end-to-end learned alternatives.
Only for evaluation, we use the 40 no-feedback and reactive sessions from the final user study as described in Sec.~\ref{sec:data_splits}, and assess performance during events where a piece enters a subgraph to which it was not previously assigned. An event is incorrect when the resulting subgraph exceeds its prescribed capacity for that piece type (i.e., a mistake). 
Applying this rule to manual annotations yielded 498 events, of which 96 incorrect and 402 correct.

We compare against several learned end-to-end baselines: a GRU~\cite{cho2014learning}, a relational Transformer~\cite{lee2019set}, and frozen MOMENT~\cite{goswami2024moment} and V-JEPA~2~\cite{assran2025v} encoders with trainable prediction heads. 
We also adopt TRAFA~\cite{mokhtar2026trafaanticipatinguseractions} by replacing its next-block rule with a capacity check over fixed, prior-derived subgraph regions. 
All models receive the same tracking stream, with V-JEPA~2 additionally receiving RGB clips.
% : MOMENT encodes tracking histories, while V-JEPA~2~\cite{assran2025v} additionally uses RGB clips with tracking-derived piece queries. 
% These comparisons cover recurrent modeling, relational attention, and time-series and video pretraining. 
% All baselines are trained on 60 pilot sessions to predict per-piece error scores, and select their thresholds on the 12 remaining pilot validation sessions. 
An anticipated mistake is evaluated as correct if it precedes the event by at least 0.44~s.
Tab.~\ref{tab:system_benchmark} shows that \tancraft{} achieves the highest precision and $F_1$, while TRAFA achieves slightly higher recall but substantially lower precision, indicating more false warnings.
% The ordering relative to \tancraft{} also holds when baselines train only on the 36 no-feedback/reactive pilot sessions, where their $F_1$ scores range from 0.33 to 0.43. These comparisons use one training run per configuration.

We also report the threshold sensitivity in Fig.~\ref{fig:system_threshold}. Increasing the threshold trades recall for precision over the plotted range, with $\gamma{=}0.55$ achieving the highest $F_1$.  
%While $\gamma{=}0.55$ was not selected based on theses results, it supports the chosen value, achieving the highest $F_1$.   
%The threshold the mean warning lead time to exceed four processing steps (approximately 0.4–0.45~s). This budget allows for one system update and a typical 200-300 ms human reaction time~\cite{card2018psychology}.

%The $\gamma$ selected pre-study on the pilot validation set achieves the highest observed $F_1$; these results were not used to select it.

% All predictions pass through the same scoring procedure. Firing frames for a piece separated by at most 15 video frames are merged into a warning episode, timestamped at its onset. Episodes raised while the piece is already incorrectly placed are excluded. 
% A warning is matched to the piece's next join and counts as timely only if it precedes that join by at least 13 video frames, approximately 0.44~s. Each join contributes at most one true or false positive; incorrect joins without timely warnings are false negatives. Warning timing accounts for logged completion timestamps for \tancraft{} and measured inference delays for the baselines. Precision, recall, and $F_1$ are pooled across sessions. 
% \tancraft{} retains its pre-study threshold $\gamma=0.55$.

\begin{table}[t]
\centering
\caption{\revised{System-level mistake anticipation benchmark}}
\label{tab:system_benchmark}
\small
\setlength{\tabcolsep}{2pt}
\begin{tabular}{@{}lcccc@{}}
\hline
& \multicolumn{3}{c}{Seen shapes}
& Unseen shapes \\
\cline{2-4}\cline{5-5}
Method & Prec. & Recall & $F_1$ & $F_1$ \\
\hline
GRU~\cite{cho2014learning}
& 0.26 & 0.61 & 0.36 & 0.33 \\
Relational Transformer~\cite{lee2019set}
& 0.24 & 0.48 & 0.32 & 0.36 \\
Frozen MOMENT~\cite{goswami2024moment}
& 0.24 & 0.57 & 0.34 & 0.37 \\
Frozen V-JEPA~2$^\dagger$~\cite{assran2025v}
& 0.28 & 0.58 & 0.38 & 0.26 \\
TRAFA (adapted)~\cite{mokhtar2026trafaanticipatinguseractions}
& 0.25 & \textbf{0.79} & 0.38 & 0.36 \\
\tancraft{}
& \textbf{0.45} & 0.77 & \textbf{0.57} & \textbf{0.51} \\
\hline
\end{tabular}

\vspace{3pt}
\begin{minipage}{\columnwidth}
\footnotesize
$^\dagger$Additional RGB input.
Unseen shapes: leave-one-shape-out evaluation.
\end{minipage}
\end{table}

\noindent\textit{Generalization to unseen shapes.}
The model used in our user study was trained on all shapes used. Here, we additionally assess system generalization to unseen shapes via a leave-one-shape-out protocol.  For each fold, we retrain the models without the held-out shape's pilot sessions, exclude that shape from validation, and evaluate on its no-feedback and reactive study sessions. Architectures and training hyperparameters remain fixed. For the held-out shape, the classifier's one-hot shape indicator is unseen during training, so subgraph assignment relies primarily on shape-independent features (piece pose, type, and cluster membership). The held-out shape's prescribed subgraph composition is provided for assembly reasoning and capacity checking. 
% Baseline thresholds are selected on the remaining pilot validation sessions, while \tancraft{} retains $\gamma=0.55$. 
We pool prediction counts across folds and compare against an all-shape control evaluated on the same study sessions.
As shown in Tab.~\ref{tab:system_benchmark}, \tancraft{} retains the highest pooled unseen-shape $F_1$ at 0.51, compared with 0.57 for its retrained all-shape control. These results support generalization to unseen target shapes within the tangram domain given their prescribed subgraph compositions.}

%-----------------------------------------------------------------------------

\subsection{User study design}
\label{sec:experimental_design}

We now turn to validating \tancraft{} via a user study, comparing two \tancraft{} variants against two baselines:
\begin{itemize}
    \item \textsc{No Feedback} does not assist the user.
    \item \textsc{Reactive Feedback} provides feedback after an incorrect placement is complete.
    \item \textsc{Predictive Feedback} warns the user when the predicted invalid-placement confidence exceeds $\gamma=0.55$.
    \item \textsc{Continuous Feedback} maps invalid-placement confidence to the loudness of a sustained tone, saturating at a confidence of $\gamma=0.65$.
\end{itemize}

Both \textsc{Predictive} and \textsc{Continuous Feedback} use the full system, differing only in how the feedback is conveyed. 

\begin{table*}[t]
\centering
\caption{User study results across feedback modes.}
\label{tab:user_study_main}
\small
\setlength{\tabcolsep}{6pt}
\begin{tabular*}{\textwidth}{@{\extracolsep{\fill}}lccccc@{}}
\hline
\textbf{Mode} &
\textbf{\# Mistakes \(\downarrow\)} &
\textbf{Overall NASA-TLX \(\downarrow\)} &
\textbf{Success Rate \(\uparrow\)} &
\textbf{Time (s) \(\downarrow\)} &
\textbf{\# Feedback} \\
\hline
\textsc{No Feedback}
& \(4.90 \pm 5.4\)
& \(3.63 \pm 1.18\)
& 0.45
& \(193.8 \pm 70.9\)
& \(0.00 \pm 0.00\) \\

\textsc{Reactive Feedback}
& \(1.95 \pm 2.4\)
& \(3.20 \pm 1.12\)
& 0.50
& \(172.9 \pm 78.4\)
& \(1.20 \pm 1.2\) \\

\textsc{Predictive Feedback}
& \(0.90 \pm 1.4\)
& \(\mathbf{3.05 \pm 1.09}\)
& \(\mathbf{0.70}\)
& \(\mathbf{148.5 \pm 77.5}\)
& \(2.75 \pm 1.8\) \\

\textsc{Continuous Feedback}
& \(\mathbf{0.75 \pm 1.0}\)
& \(3.32 \pm 1.09\)
& \(\mathbf{0.70}\)
& \(170.1 \pm 64.7\)
& \(1.85 \pm 1.7\) \\
\hline
\end{tabular*}

\vspace{2pt}
\footnotesize
Values are mean \(\pm\) standard deviation except success rate, which is reported as a proportion. Overall NASA-TLX is the unweighted mean of the six NASA-TLX subscales, with perceived success scores reversed so that lower values indicate lower workload. One effort rating was missing, so Overall NASA-TLX is computed over \(N=19\); all other metrics use \(N=20\).
\end{table*}

\revised{The user study comprises 80 sessions from 20 new participants recruited from %(anonymized for review)
university campus, disjoint from the pilot cohort, with 20 sessions per feedback condition. As a result, no user study sessions were used for model training or tuning.}
None were affiliated with the research presented in this paper, and all provided informed consent. 
% Participants assembled the tangram targets from seven colored pieces (Fig.~\ref{fig:tang_piece}) on a tabletop workspace while the system used the overhead RGB stream for online inference. 
Participants also wore a head-mounted camera that recorded an egocentric view for dataset release only; this stream was not used by the system. A monitor displayed the target silhouette and a color-coded reference panel indicating the semantic subgraph associated with each piece, without revealing the final piece placements as shown in Fig.~\ref{fig:teaser}. 

\begin{figure*}[t]
\centering
\includegraphics[width=\textwidth]{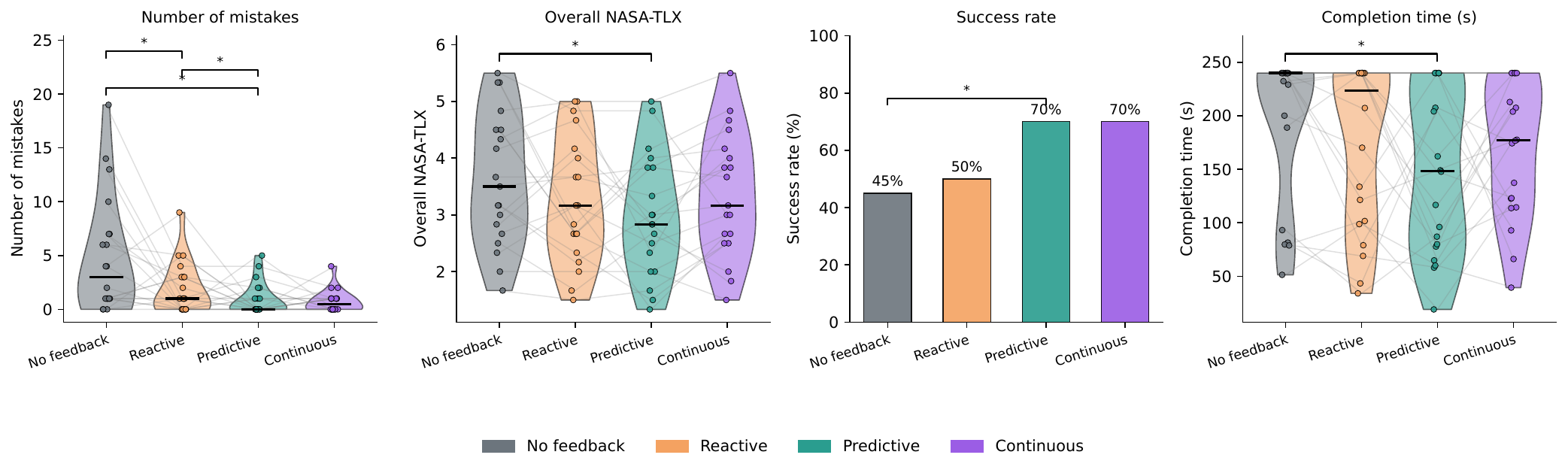}
\caption{\textbf{Within-subject comparison of feedback policies.} Each panel shows
participant-level outcomes for all four modes evaluated. Lines connect results for the same participants. Asterisks mark
significant planned model-based contrasts (\(p<.05\)).}
\label{fig:user_study_violin}
\end{figure*}

After written instructions and one warm-up round on the \textit{house} shape, each participant completed four main rounds on \textit{fox}, \textit{swan}, \textit{camel}, and \textit{penguin}. Each round ended upon completion and was capped at four minutes. Participants completed the NASA-TLX questionnaire~\cite{hart1988development} to assess their perceived workload immediately after each main round.
Feedback order was counterbalanced with a \(4\times4\) Williams Latin square replicated five times, and mode-target pairing was balanced so that each \((\textit{mode},\textit{shape})\) pair occurred five times. The warm-up mode was assigned by blocked randomization and differed from the first main-round mode. To reduce memorization and order effects, the initial piece-color arrangement and the color order in the reference panel were re-randomized at the start of every main round.

We evaluated the feedback conditions by their task success rate, number of mistakes, completion time, number of feedback events, and perceived workload. Our analysis focuses on the three primary policies: \textsc{No Feedback}, \textsc{Reactive Feedback}, and \textsc{Predictive Feedback}. \textsc{Continuous Feedback} is included as an additional comparison to test whether continuously providing feedback improves or burdens the interaction. Because the metrics are counts (mistakes), binary (success), continuous (TLX), and censored (completion time, capped at 240 s), we analyze each with an appropriate mixed-effects model with modes as a fixed effect and participants as a random effect: Poisson for mistakes, binomial for success, linear for TLX, and a Cox proportional-hazards model treating capped trials as right-censored for completion time. We report planned contrasts via estimated marginal means with Bonferroni correction, using one-sided tests for our directional hypotheses. 
%-----------------------------------------------------------------------------

\subsection{User Study Results}
\label{sec:user_study_results}

We show the results of our user study in Tab.~\ref{tab:user_study_main}, while Fig.~\ref{fig:user_study_violin} visualizes the participant-level distributions. Participants made \(0.90\pm1.4\) mistakes with \textsc{Predictive Feedback}, compared with \(4.90\pm5.4\) mistakes with \textsc{No Feedback} and \(1.95\pm2.4\) mistakes with \textsc{Reactive Feedback}, resulting in a significant reduction relative to both \textsc{No Feedback} (\(p<.001\)) and \textsc{Reactive Feedback} (\(p=.036\)). 
% \textsc{Reactive Feedback} also reduced mistakes relative to no feedback (\(p<.001\)). 
Perceived workload followed a similar pattern. \textsc{Predictive Feedback} produced the lowest workload score (\(3.05\pm1.09\)), compared to \textsc{No Feedback} (\(3.63\pm1.18\)) and \textsc{Reactive Feedback} (\(3.20\pm1.12\)), with the reduction relative to \textsc{No Feedback} significant (\(p=.015\)). 
% Thus, predictive feedback did not trade fewer mistakes for higher subjective burden; instead, it reduced mistakes while also lowering the subjective workload. 
While \textsc{Predictive Feedback} does produce more warning events than \textsc{Reactive Feedback} (\(2.75\pm1.8\) vs.\ \(1.20\pm1.2\)), the absolute number remained low.

The other performance metrics show the same favorable direction. Success rate increased from \(0.45\) with \textsc{No Feedback} and \(0.50\) with \textsc{Reactive Feedback} to \(0.70\) with \textsc{Predictive Feedback} (\(p=.029\) vs.\ \textsc{No Feedback}). Completion time also decreased: \textsc{Predictive Feedback} produced the shortest mean completion time (\(148.5\pm77.5\) s), and the Cox model confirmed faster completion than \textsc{No Feedback} (\(p=.002\)). \textsc{Continuous Feedback} achieved the lowest mean mistake count and matched \textsc{Predictive Feedback}'s success rate, but had higher mean workload and longer mean completion time.

%%%%%%%%%%%%%%%%%%%%%%%%%%%%%%%%%%%%%%%%%%%%%%%%%%%%%%%%%%%%%%%%%%%%%%%%%%%%%%%%
\section{Discussion}
\label{sec:discussion}

\revised{A central challenge in non-sequential assembly is distinguishing valid exploration from an impending mistake without prescribing the user's next action. \tancraft{} addresses this by evaluating predicted placements against the evolving assembly's remaining capacities, allowing different construction orders while selectively warning about incompatible outcomes. The system-level benchmark shows stronger mistake anticipation than the evaluated baselines, including on unseen tangram shapes. The user study extends prior findings on proactive assistance to error warnings in non-sequential assembly, showing that selective predictive feedback reduces mistakes relative to no feedback and reactive warnings.}

Moreover, \textsc{Continuous Feedback} offers an alternative to binary predictive warnings by conveying invalid-placement confidence through a graded tone. In our study, it achieved similarly low mistake counts and the same success rate as predictive feedback, suggesting that graded feedback can also support error prevention. 
However, continuous feedback was associated with higher mean workload and longer mean completion time.
% , suggesting a potential interaction cost of continuously conveying confidence. 
Adaptive gain, confidence smoothing, or user-specific thresholds may help reduce this burden. 

Although tangram is a controlled domain, it captures key challenges of non-sequential assembly: objective correctness, flexible construction order, and exploratory manipulation. \revised{Moreover, the leave-one-shape-out evaluation supports generalization to unseen tangram targets when their subgraph compositions are provided. However, this offline evaluation does not establish generalization to new piece inventories, workspaces, or assembly tasks.} Extending the framework to more complex assemblies and robot-assisted manipulation remains future work.

Beyond system evaluation, we will release our multi-view tangram assembly dataset comprising \revised{$152$ sessions} and eight hours of activity, recorded synchronously from overhead and egocentric cameras (${\approx}16$ hours of video in total). The dataset includes tracked objects and hands, subgraph assignments, placements, mistakes, and feedback metadata. By capturing free-form manipulation and user responses to feedback, it can support future research on action forecasting, mistake anticipation, assistance timing, and reasoning tasks.

%%%%%%%%%%%%%%%%%%%%%%%%%%%%%%%%%%%%%%%%%%%%%%%%%%%%%%%%%%%%%%%%%%%%%%%%%%%%%%%%
\section{Conclusion}
\label{sec:conclusion}

We presented \tancraft{}, a real-time predictive assistance framework for non-sequential assembly that tracks the scene, reasons about the partial assembly, forecasts active-piece motion, and selectively intervenes before incompatible placements. \revised{System-level comparisons show improved mistake anticipation over the evaluated learned baselines, including generalization to unseen tangram shapes.}
In a within-subject study with $20$ participants, our system significantly reduced the number of mistakes,
% from $4.90$ without feedback and $1.95$ with reactive feedback to $0.90$
while achieving the lowest mean perceived workload and shortest mean completion time. 
% These findings highlight the value of combining assembly-aware error anticipation with selective intervention to improve human performance. 
More broadly, \tancraft{} provides a step toward cost-aware assistance in automation and human-robot workcells, where warnings should be both reliable and timely.

%%%%%%%%%%%%%%%%%%%%%%%%%%%%%%%%%%%%%%%%%%%%%%%%%%%%%%%%%%%%%%%%%%%%%%%%%%%%%%%%
\addtolength{\textheight}{-12cm}   % This command serves to balance the column lengths

%%%%%%%%%%%%%%%%%%%%%%%%%%%%%%%%%%%%%%%%%%%%%%%%%%%%%%%%%%%%%%%%%%%%%%%%%%%%%%%%



% \section*{ACKNOWLEDGMENT}

\bibliographystyle{IEEEtran}
\bibliography{references}
\end{document}